\documentclass[manuscript]{acmart}

\AtBeginDocument{%
  \providecommand\BibTeX{{%
    \normalfont B\kern-0.5em{\scshape i\kern-0.25em b}\kern-0.8em\TeX}}}

\newcommand\blfootnote[1]{%
  \begingroup
  \renewcommand\thefootnote{}\footnote{#1}%
  \addtocounter{footnote}{-1}%
  \endgroup
}

\setcopyright{acmcopyright}
\copyrightyear{2021}
\acmYear{2021}
\setcopyright{rightsretained}
\acmConference[RecSys '21]{Fifteenth ACM Conference on Recommender Systems}{September 27-October 1, 2021}{Amsterdam, Netherlands}
\acmBooktitle{Fifteenth ACM Conference on Recommender Systems (RecSys '21), September 27-October 1, 2021, Amsterdam, Netherlands}
\acmDOI{10.1145/3460231.3474605}
\acmISBN{978-1-4503-8458-2/21/09}

\begin{document}

\title{Scaling TensorFlow to 300 million predictions per second}

\author{Jan Hartman}
\email{jhartman@outbrain.com}
\author{Davorin Kopi\v{c}}
\email{dkopic@outbrain.com}
\affiliation{
  \institution{Zemanta, an Outbrain company}
  \country{Slovenia}
}

\blfootnote{Authors' emails: jhartman@outbrain.com; dkopic@outbrain.com.}

\begin{abstract}
We present the process of transitioning machine learning models to the TensorFlow framework at a large scale in an online advertising ecosystem. In this talk we address the key challenges we faced and describe how we successfully tackled them; notably, implementing the models in TF and serving them efficiently with low latency using various optimization techniques.
\end{abstract}




\maketitle

\section{Introduction}

In this work, we describe the process of scaling machine learning models implemented in the TensorFlow machine learning framework to over 300 million predictions per second at Zemanta, an Outbrain company. Zemanta is a demand-side platform (DSP) in the real-time bidding (RTB) ecosystem, which is a fast-growing part of online advertising~\cite{wang2016displayrtb}. In RTB, several DSPs (bidders) compete for advertising space online by bidding for it in real-time while a web page is still loading. The advertising space is sold on a per-ad impression basis, which enables selling virtual ad space at market value. Through the use of machine learning, RTB also enables advertisers to maximize their KPIs such as click-through rate (CTR). Estimating the CTR of ads is therefore one of the central problems in RTB since it allows advertisers to only bid and pay for measurable user responses, such as clicks on ads. Having a good click prediction model is thus of significant importance. Many new algorithms and modeling techniques are proposed each year by researchers in academia and industry~\cite{mcmahan2013ad, he2014practical}.

The RTB field comes with a few intrinsic properties: large amounts of data (Zemanta receives over a million bid requests per second) and a low latency requirement (the auctions have a maximal allowed response time of 100 milliseconds). The distribution of the data also changes rapidly, meaning that models need to be updated with new data frequently in order to stay competitive. All of this means that we are heavily constrained in our choice of models to those that enable quick incremental updates. Because of that, we initially used custom-written logistic regression and factorization machines (FM)~\cite{rendle2010factorization} based models implemented in Golang~\cite{donovan2015go}. They were limited by their expressiveness and the need to manually implement all learning procedures, slowing down experimentation and limiting their predictive performance. These factors contributed to the decision to adopt the TensorFlow framework and replace existing models with more expressive ones.

\section{Motivation}
TensorFlow (TF)~\cite{abadi2016tensorflow}, developed by Google Brain, is currently one of the most widely used machine learning frameworks. It facilitates much quicker research and decreases the time to production for new models. It also standardizes our approach to an open-source framework that is well known in both industry and academia. This in turn enables more knowledge sharing and requires less time to learn an in-house framework for new data scientists. TF's major downside is additional complexity, both in modeling and production usage. 
We first implemented FMs in TF to validate the correctness of the new pipeline, then moved to a replacement for FMs -- DeepFM~\cite{guo2017deepfm}, a neural network-enhanced FM. The reason for choosing a much more complex model was to significantly outperform simpler FM models, bring lifts in business metrics and utilize TF’s potential well. Using the same FM model would bring no increases in predictive performance and would not offset the complexity costs.

\section{Challenges}
Due to our very specific and demanding use case, we faced a number of challenges when implementing and scaling new models in TF. We divide them into the implementation, serving, and optimization sections.

Our bidder is a monolith, which is beneficial in three areas: ease of deployment, latency, and engineering costs. A series of services would have higher and/or less predictable latency. This would be a major downside in our system -- every millisecond saved on service network calls can instead be used for actual processing and better ML models. For this reason, we chose to use the core TF framework inside the bidder application (we use the Golang wrapper\footnote{\url{https://pkg.go.dev/github.com/tensorflow/tensorflow/tensorflow/go}}). An alternative is TensorFlow Serving\footnote{\url{https://www.tensorflow.org/tfx/guide/serving}}, which is a premade service for serving TF models with additional features such as batching capabilities. Due to the above-mentioned reasons, we avoided it.

We additionally do not utilize GPUs for inference in production. At our scale, outfitting each machine with one or more top-class GPUs would be prohibitively expensive, and on the other hand, having only a small cluster of GPU machines would force us to transition to a service-based architecture. Given that neither option is particularly preferable and that our models are relatively small compared to state-of-the-art models in other areas of deep learning (such as computer vision or natural language processing), we consider our approach much more economical. Our use case is also not a good fit for GPU workloads due to our models using sparse weights.

\subsection{Implementation}
To implement an effective training loop for TF models, we researched, implemented, and tested a variety of approaches. Case studies of high-throughput online training and serving in TF are scarce and the documentation is often not specific enough, which forced us to read through the source code and benchmark prototypes, recognizing pitfalls in the process.

TF offers a massive ecosystem and plenty of libraries with state-of-the-art algorithm implementations. It is very easy to pick a feature-rich off-the-shelf implementation, however, we found that these are mostly unoptimized. 
We then decided to implement the algorithms ourselves, but even starting was not trivial: TF has APIs of varying levels of abstraction, from the most easy-to-use, but often inefficient (Estimator API) to the most low-level operations. We chose Keras\footnote{\url{https://keras.io/}} as it is a thin wrapper around low-level TF operations and maintains a high level of performance while being easy to understand.
Since TF is a very feature-rich and resource-heavy library, we also had to consider how much of our machine learning pipeline we would implement in it. We opted to set aside feature transformation and interaction for now and only implement learning algorithms -- they are the smallest part that can be replaced yet offer the highest potential for improvement.

Because the Golang TF wrapper supports only prediction, we had to implement the training loop in Python. The script is connected to our Golang data pipeline through its standard input as a subprocess. Data is sent to it in a highly efficient binary format, requiring no parsing -- this was a ~25\% improvement in speed over a CSV format. 
The data is then read in a background thread to prevent the model from being idle while waiting for data. With this, we managed to retain a high throughput through the entire training pipeline, having only the model as a potential bottleneck. Efficient inputs and outputs proved to be key for low-latency prediction as well, where we significantly decreased the time spent on expensive serialization and copying of input data by joining all input features into a single tensor.

\subsection{Serving}
We found that by using the Golang TF wrapper out of the box, the DeepFM models incurred a much higher CPU usage due to the compute-intensive neural networks. Despite bringing a significant lift in business metrics, scaling this approach to 100\% of our traffic would have required a significant hardware investment. As the entire world was (and still is) facing a shortage of microchips, this would prove difficult, not to mention expensive.

To combat this, we saw a need to make these computations less expensive. Reducing the model's neural network size was to be avoided if possible as it would also reduce the model's predictive performance. By diving deeply into TF, we realized that the computation is far more efficient (per example) if we increase the number of examples in a compute batch. This low-linear growth is due to TF code being highly vectorized.  TF also has some overhead for each compute call, which is then amortized over larger batches. Given this, we figured that in order to decrease the number of compute calls, we needed to join many requests into a single computation.

We built an autobatching system contained entirely within a running bidder instance, avoiding network calls. Since each instance receives thousands of bid requests per second, we can reliably join the computations from many requests, creating bigger batches. We did this by having a few batcher threads which receive data from incoming requests, create batches and initialize computation once the batch has been filled. The computation is always initialized at least every few milliseconds to prevent timeouts since it is possible that the batch isn't filled in this time window.
This implementation is highly optimized and is able to decrease the number of compute calls by a factor of 5, halving the CPU usage of TF compute. In rare cases that a batcher thread does not get CPU time, those requests will time out. However, this happens on fewer than 0.01\% of requests. We observed a slight increase in the average latency -- by around 5 ms on average, which can be higher in peak traffic. We put SLAs and appropriate monitoring into place to ensure stable latencies. As we did not increase the percentage of timeouts substantially, this was highly beneficial and is still the core of our TF serving mechanisms.

\subsection{Optimizations}

The models we implemented in TF were initially much slower than the custom-built FMs. To find potential speedups, we heavily utilized the inbuilt TF profiler to find the operations which take the longest to execute. We were able to create many possible improvements with these insights, the most common being various redundant reshape or transform operations. One of the more interesting findings was discovering that the Adam optimizer was much slower than Adagrad (around ~50\%), despite the difference in the number of operations being small. 
The profiler showed that gradient updates on our sparse weights require a large amount of computational time. This is due to the model's weights being sparse (the features are largely categorical and thus very sparse) and the optimizer not taking this fact into account. Since replacing Adam with Adagrad meant a significant deterioration of the deep models' performance, we looked for other solutions. Switching to the Lazy Adam optimizer\footnote{\url{https://www.tensorflow.org/addons/api_docs/python/tfa/optimizers/LazyAdam}} proved very beneficial as it handles sparse weights very efficiently. We found that it sped up overall training over 40\%, bringing it up to par with Adagrad in this regard.

In RTB, the data distribution changes rapidly, presenting a need for continuous model training. This is why we continuously update our models with a job and deploy the trained models onto the fleet of bidder machines. Because we run many models in production at the same time, the memory and storage requirements are significant.
Since we use adaptive optimizers such as Adam, this also requires storing the weights' moments and variances -- instead of one, three values are stored for each parameter, increasing the saved model size threefold. However, these values are not actually needed for prediction, only for training. We utilized this to construct an optimization routine that strips the model of these values, reducing the amount of data that is pulled to our bidder machines by 66\% and decreasing the memory usage and costs.

\section{Conclusion}
In this extended abstract, we described the process of transitioning machine learning models to the TensorFlow framework and serving them at a large scale. The key challenges we faced in our use case were related to compute resources, prediction latency, and training throughput. By implementing autobatching in serving, we halved TF's CPU usage and retained acceptable latencies. Along with thoroughly understanding the models, we also put effort into putting together an efficient training pipeline: using a binary data format instead of CSV, utilizing the TF profiler to remove bottlenecks in the models, and using the correct optimizer have all brought significant speedups. We have also implemented many other smaller and more specific optimizations such as stripping the optimizer weights to reduce the saved model size. 
Overall, using TF has brought significant lifts in business metrics and vastly increased the speed of research. To make the best use of it, we are continuing to optimize our pipelines and serving stack.

\section*{Speaker Bio}

\textbf{Jan Hartman} is a data scientist/machine learning engineer at Zemanta, where he works with high-throughput, low-latency machine learning pipelines at a large scale. He leads the initiative for implementing and applying state-of-the-art models for click prediction. Before joining Zemanta, he worked on research projects in the fields of distributed computing, neural network optimization, and cryptography. 
Honors BSc degree in Computer \& Information Science from the University of Ljubljana, currently finishing MSc in Data Science. Open-source contributor. Research interests include deep learning, neural network embeddings, and model compression. 
\newline

\noindent
\textbf{Davorin Kopi\v{c}} is the Head of Data Science at Zemanta, an Outbrain company. He is leading the data science research and development department working on high throughput and scalable applied machine learning algorithms for real-time bidding on online advertising space, autonomous campaign optimization algorithms, automated fraud detection, large-scale data analysis and similar. His expertise are in online advertising technology sector, scalable machine learning systems and moving machine learning from research to production. Computer Science and Artificial Intelligence (BSc, First Class Honours) graduate from the University of Sussex in Brighton, UK. President of the Machine Learning Society at University of Sussex (2013-2015).

\bibliographystyle{ACM-Reference-Format}
\bibliography{bibliography}

\end{document}